\documentclass{article} 
\usepackage{iclr2024_conference,times}

\usepackage{amsmath,amsfonts,bm}

\def\eqref#1{equation~\ref{#1}}

\def\1{\bm{1}}

\def\rvx{{\mathbf{x}}}

\def\rvz{{\mathbf{z}}}

\def\vg{{\bm{g}}}

\def\vm{{\bm{m}}}

\def\vx{{\bm{x}}}

\def\vz{{\bm{z}}}

\def\mI{{\bm{I}}}

\def\mW{{\bm{W}}}
\def\mX{{\bm{X}}}

\def\mZ{{\bm{Z}}}

\DeclareMathAlphabet{\mathsfit}{\encodingdefault}{\sfdefault}{m}{sl}
\SetMathAlphabet{\mathsfit}{bold}{\encodingdefault}{\sfdefault}{bx}{n}

\newcommand{\E}{\mathbb{E}}

\newcommand{\R}{\mathbb{R}}

\newcommand{\KL}{D_{\mathrm{KL}}}

\usepackage{hyperref}
\usepackage{url}
\usepackage{algorithm}
\usepackage{algorithmic}
\newcommand{\pluseq}{\mathrel{+}=}
\usepackage{graphicx}
\usepackage{booktabs} 
\usepackage{siunitx} 

\usepackage{tabularray}

\title{Sample as you Infer: \\ Predictive Coding with Langevin Dynamics}

\iclrfinalcopy

\author{Umais Zahid \thanks{Correspondence to umaiszahid@outlook.com}  \\
Huawei Technologies R\&D\\
London, UK
\And
Qinghai Guo  \\
Huawei Technologies R\&D\\
London, UK \\
\And
Zafeirios Fountas  \\
Huawei Technologies R\&D\\
London, UK 
}

\iclrfinalcopy 
\begin{document}

\maketitle

\begin{abstract}
We present Langevin Predictive Coding (LPC), a novel algorithm for deep generative model learning that builds upon the predictive coding framework of computational neuroscience. By injecting Gaussian noise into the predictive coding inference procedure and incorporating an encoder network initialization, we reframe the approach as an amortized Langevin sampling method for optimizing a tight variational lower bound. To increase robustness to sampling step size, we present a lightweight preconditioning technique inspired by Riemannian Langevin methods and adaptive SGD. We compare LPC against VAEs by training generative models on benchmark datasets. Experiments demonstrate superior sample quality and faster convergence for LPC in a fraction of SGD training iterations, while matching or exceeding VAE performance across key metrics like FID, diversity and coverage.
\end{abstract}

\section{Introduction}

In recent decades the Bayesian brain hypothesis has emerged as a compelling general framework for understanding perception and learning in the brain \citep{pouget_probabilistic_2013, clark_whatever_2013, kanai_cerebral_2015}. Under this framework, the brain is posited as encoding a probabilistic generative model engaged in a joint scheme of inference over the hidden causes of its observations and learning over its model parameters. One of the most popular instantiations of this view is predictive coding (PC), a computational scheme which employs hierarchical latent Gaussian generative models with complex, non-linear conditional parameterizations. In recent years, PC has garnered substantial attention for its potential to elucidate cortical function \citep{rao_predictive_1999, friston_does_2018, mumford_computational_1992, hosoya_dynamic_2005, hohwy_predictive_2008, bastos_canonical_2012, shipp_neural_2016, feldman_attention_2010, fountas_predictive_2022}. Despite its predictive appeal in the cognitive sciences, the practical applicability and performance of PC in training deep generative models, akin to those conjectured to operate in the brain, has yet to be fully realized \citep{zahid_predictive_2023}.

Concurrent to these developments in the cognitive sciences, a separate revolution has been occurring in the statistical literature driven by the use of gradient-based Monte Carlo sampling methods such as Hamiltonian Monte Carlo (HMC) \citep{roberts_exponential_1996, neal_mcmc_2011, hoffman_no-u-turn_2011, girolami_riemann_2011, ma_is_2019}. These methods facilitate the sampling of intractable distributions through the intelligent construction of Markov chains with proposals informed by gradient information from the log density being sampled. Notably, one of the simplest algorithms within this class is the overdamped Langevin algorithm \citep{rossky_brownian_1978, roberts_exponential_1996, roberts_optimal_1998}, which admits an interpretation as both a limiting case of HMC, and as a discretisation of a Langevin diffusion \citep{neal_mcmc_2011}.        

This paper introduces several advancements aimed at extending the PC framework using techniques from gradient-based Markov Chain Monte Carlo (MCMC) for use in training deep generative models:
\begin{itemize}
    \item We show that by injecting appropriately scaled Gaussian noise, the standard PC inference procedure may be interpreted as an (unadjusted) overdamped Langevin sampling. 
    \item Utilizing these Langevin samples, we compute gradients with respect to a tight evidence lower bound (ELBO), which model parameters may be optimised against.
    \item To improve chain mixing time, we train approximate inference networks for amortized warm-starts and evaluate three distinct objectives for their optimization.
    \item We investigate and validate a light-weight diagonal preconditioning strategy for increasing robustness to the Langevin step size, inspired by adaptive optimization techniques.
\end{itemize}

\section{Methodology}
\subsection{Inference as Langevin Dynamics}

The standard PC recipe for inference and learning under a generative model, for static observations, may be described succinctly as follows \citep{rao_predictive_1999, bogacz_tutorial_2017, millidge_predictive_2020}:

\begin{enumerate}
    \item Define a (possibly hierarchical) graphical model over latent ($\rvz \in \R^d$) and observed ($\rvx \in \R^n$) states with parameters $\pmb\theta$: $\log p(\rvx,\rvz|\pmb\theta)$
    \item For each observation $\vx^{(i)} \sim \mathcal{D}$, where $\mathcal{D}$ is the data-generating distribution. \\
        \textbf{Inference: } 
        Iteratively enact a gradient ascent on $\log p(\vx^{(i)},\vz|\theta)$ with respect to latent states ($\vz$)
            \begin{align}
                \vz^{(t)} = \vz^{(t-1)} + \gamma \nabla_{\vz} \log p(\rvx^{(i)}, \rvz^{(t-1)}|\pmb\theta) \label{eq:standard_pc_iteration}
            \end{align}
            Until you obtain an MAP estimate: \\ $\vz_{\text{MAP}} = \max_{\vz}  \log p(\rvx^{(i)}, \rvz|\pmb\theta)$
            
        \textbf{Learning: } Update model parameters $\theta$ using stochastic gradient descent with respect to the log joint evaluated at the MAP (averaged over multiple observations if using mini-batches): 
        \begin{align}
            \pmb\theta^{(i)} = \pmb\theta^{(i-1)} + \alpha \nabla_{\pmb\theta} \log p(\vx^{(i)}, \vz_{{\text{MAP}}}|\pmb\theta^{(i-1)}) \label{eq:standard_pc_learning}
        \end{align}
\end{enumerate}

One simple and relevant framing of this process is that of a variational ELBO maximising scheme under the assumption of a Dirac delta (point-mass) approximate posterior \citep{friston_learning_2003, friston_theory_2005, friston_predictive_2009, zahidCurvatureSensitivePredictiveCoding}. In practice, the restrictiveness of this Dirac delta posterior significantly impairs the quality of the resultant model due to the expected divergence between the true model posterior and the Dirac delta function situated at the MAP estimate. Indeed, previous attempts at reducing the severity of this assumption, by adopting quadratic approximations to the posterior at the MAP, \citep{zahidCurvatureSensitivePredictiveCoding}, succeeded in improving model quality to a degree, but suffered from high computational cost while still performing significantly worse than their variational auto-encoder counterparts.

Our contribution begins with the observation that by injecting appropriately scaled Gaussian noise into Equation \ref{eq:standard_pc_iteration}, one obtains an unadjusted Langevin algorithm (ULA). Specifically, the ULA may be considered the discretisation of a continuous-time Langevin diffusion \citep{rossky_brownian_1978, roberts_exponential_1996}, characterised by the following stochastic differential equation,
\begin{figure}[h]
\begin{center}
\includegraphics[width=0.7\textwidth]{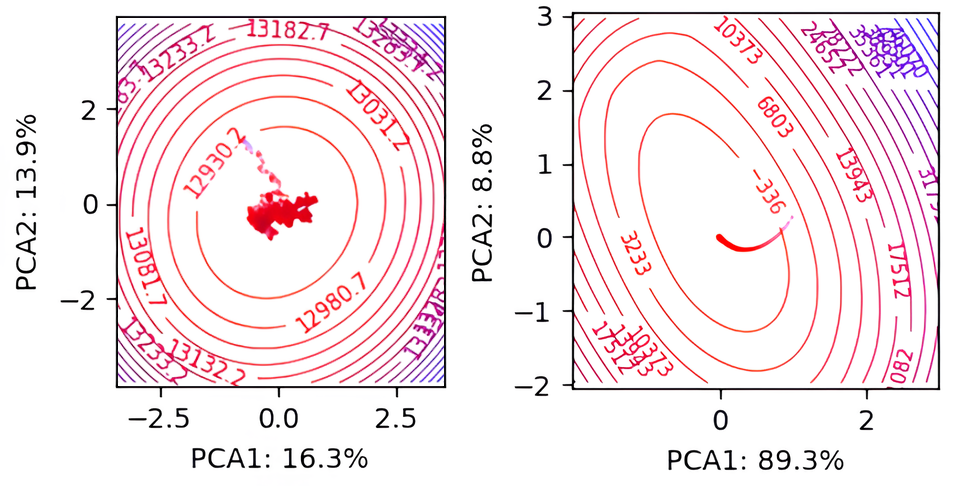}
\label{fig:low_dimensional}
\end{center}
\caption{Projection of high-dimensional latent state trajectories under standard PC inference (right), and Langevin PC sampling (left), using normalised PCA trajectories. Latent state dynamics under Langevin PC result in a principled exploration of the posterior. More examples trajectories, and further details on how these were computed may be found in Appendix \ref{appendix:pca_latent_trajectory}. Contour lines and hue correspond to values of the negative log joint probability (blue high, red low), marker brightness corresponds to time-step (earlier is lighter).} 
\end{figure}
\begin{align}
    d\mZ_t &= -\nabla_{\vz} U(\mZ_t)dt + \sqrt{2}d\mW_t
    \intertext{where $\mW_t$ is a d-dimensional Brownian motion and admits a unique invariant density equal to $\frac{e^{-U(\vz)}}{\int_{\R^d} e^{-U(\vz)} dz}$ under mild conditions. Setting the potential energy ($U(\vz)$) to $-\log p(\rvx^{(i)}, \rvz|\pmb\theta)$, for an observation $\rvx^{(i)}$ gives us:}
    d\mZ_t &= \nabla_{\vz} \log p(\rvx^{(i)}, \mZ_t|\pmb\theta)dt + \sqrt{2}d\mW_t
\end{align}
for which the corresponding Euler–Maruyama discretisation scheme is:
\begin{align}
    \vz^{(t)} &= \vz^{(t-1)} + \gamma \nabla_{\vz} \log p(\vx^{(i)}, \vz^{(t-1)}|\pmb\theta) + \sqrt{2\gamma}\eta \label{eq:discretisation}
\end{align}

with $\eta \sim \mathcal{N}(\pmb0, \mI)$. This is simply equal to a standard PC inference iteration (Equation \ref{eq:standard_pc_iteration}) with the addition of some scaled Gaussian noise.
With the inclusion of this Gaussian noise, the resultant iterates $\vz^{(t)}$ would thus be interpretable as samples of the true model posterior, as $t \to \infty$, up to a bias induced by discretisation \citep{besagDiscussion, roberts_exponential_1996}.

Next, we note that by treating the (biased) samples from our Langevin chain as samples from an approximate posterior instead, we may compute gradients of a Monte Carlo estimate for the evidence lower-bound with respect to our model parameters $\pmb\theta$: 
\begin{align}
\nabla_{\pmb\theta} \mathcal{L}_{\text{ELBO}} &= 
    \nabla_{\pmb\theta}\left[ \E_{\tilde{p}(\rvz|\rvx)}[\log p(\rvx,\rvz|\pmb\theta) - \log \tilde{p}(\rvz|\rvx) ] \right] \\
    \intertext{Where the approximate posterior $\tilde{p}(\rvz|\rvx)$ corresponds to the empirical distribution of our Langevin chain. Because we are only interested in gradients with respect to our parameters $\pmb\theta$, the intractable entropy term of our sample distribution may be ignored}
    &= \nabla_{\pmb\theta}\left[ \E_{\tilde{p}(\rvz|\rvx)}[\log p(\rvx,\rvz|\pmb\theta)]\right] \nonumber \\
    & \quad - \underbrace{\nabla_{\pmb\theta}\left[\E_{\tilde{p}(\rvz|\rvx)}[\log \tilde{p}(\rvz|\rvx)] \right]}_{=0} \\
    &\approx \nabla_{\pmb\theta} \frac{1}{T} \sum_t \log p(\vx,\vz^{(t)}|\pmb\theta)
\end{align}
Crucially, optimisation of this ELBO simply requires computing the gradient of our negative potential energy $\log p(\vx, \vz|\pmb\theta)$, with respect to $\pmb\theta$ rather than $\vz$, and is (computationally) identical to the learning step in Equation \ref{eq:standard_pc_learning}. From the perspective of neurobiological plausibility, this result is a pleasant surprise, as there already exists a substantial literature on how the dynamics described by Equation \ref{eq:standard_pc_iteration} and \ref{eq:standard_pc_learning} may be implemented neuronally \citep{friston_learning_2003, friston_theory_2005, shipp_neural_2016, bastos_canonical_2012}. Thus, the Langevin PC algorithm demands no additional neurobiological machinery other than the injection of Gaussian noise into our standard PC iterates. We briefly discuss the possible implications of this in Section \ref{sec:conclusion}. 

From the perspective of an in-silico implementation, these gradients may be collected iteratively as the Markov chain is constructed, resulting in constant memory requirements independent of the chain length $T$, while reusing portions of the same backward pass used to compute our Langevin drift: $\nabla_{\vz}\log p(\vx, \vz|\pmb\theta)$. 
\subsection{Amortised Warm-Starts}

It is well-known that MCMC sampling methods, while powerful in theory, are notoriously sensitive to their choice of hyperparameters in practice \citep{steve_brooks_andrew_gelman_galin_jones_xiao-li_meng_handbook_2011}. One such choice is the state of initialisation for a Markov chain. A poor initialisation, far from the typical set of the invariant density will result in an inefficient chain with poor mixing time. This is of particular importance if we require constructing this Markov chain within each SGD training iteration. Traditional strategies to ameliorate this issue generally appeal to burn-in, i.e the discarding of a series of initial samples \citep{andrew_gelman_bayesian_2015}, or by initialising at the MAP found via numerical optimisation \citep{salvatier_probabilistic_2015}. Such strategies are costly, particularly for our Langevin dynamics as they require expensive and wasted network evaluations. 

We resolve this issue by training an amortised warm-up model (equivalently, an approximate inference model) conditional on observations. This allows us to provide a warm-start to our Langevin chain that is ideally within the typical set. In the context of the computational neuroscience origins of predictive coding, this formulations appeal compellingly to a dichotomy frequently identified in computational neuroscience. Namely, between fast but approximate feed-forward perception, vs slower but precise recurrent processing. The joint occurence of which, within the visual cortex in particular, has long been noted for it's importance in object recognition \citep{lamme_distinct_2000, mohsenzadeh_ultra-rapid_2018, kar_evidence_2019}. 

Architecturally this network may be chosen to resemble standard encoders, in encoder-decoder frameworks such as the VAE \citep{kingma_auto-encoding_2014}, however the availability of (biased) samples from the model posterior obtained through Langevin dynamics afford us greater flexibility in how we train it. Here we propose and validate three objectives for training our amortised warm-start model: the forward KL, reverse KL and Jeffrey's divergence.  

\subsubsection{Forward KL}
Given Langevin samples from the model posterior, the most obvious objective for optimising our approximate inference network is the expected forward Kullback–Leibler divergence between the model posterior and our approximate posterior, with expectation approximated with mini-batches of observations. Specifically, the forward KL divergence can be separated into an intractable but encoder-independent entropy term, and a cross entropy term for which we may obtain a Monte Carlo estimate using our Langevin samples:
\begin{align}
    \KL(\tilde{p}(\rvz|\rvx)\vert q(\rvz|\rvx,\pmb\phi)) &= \E_{(\tilde{p}(\rvz|\rvx)}\left[\log \frac{\tilde{p}(\rvz|\rvx)}{q(\rvz|\rvx,\pmb\phi)}\right]
\end{align}
where we are exclusively interested in obtaining gradients with respect to $\pmb\phi$, and as such:
\begin{align}
\!\!\!\!\!\!\nabla_{\pmb\phi} \KL(\tilde{p}(\rvz|\rvx) \vert q(\rvz|\rvx,\pmb\phi)) &= - \nabla_{\pmb\phi} \E_{\tilde{p}(\rvz|\rvx)}\left[\log q(\rvz|\rvx,\pmb\phi)\right] \nonumber \\
&\quad\!\! +  \underbrace{\nabla_{\pmb\phi} \E_{\tilde{p}(\rvz|\rvx)}\left[\log \tilde{p}(\rvz|\rvx)\right]}_{0} 
\\
&\!\!\!\!\!\!\!\!\!\!\!\!\!  = - \nabla_{\pmb\phi} \E_{\tilde{p}(\rvz|\rvx)}\left[\log q(\rvz|\rvx,\pmb\phi)\right] \label{eq:forward_kl}
\end{align}
which is simply the cross-entropy between our empirical Langevin posterior distribution and our approximate inference model. We will denote the Monte Carlo estimate for this approximate inference objective for a mini-batch of observations and a single batch of their associated posterior samples, as $\mathcal{L}_{A_F}(\vx, \vz)$.

\subsubsection{Reverse KL}
While the forward KL is readily available given our access to samples from the posterior, its well-known moment matching behaviour may result in an initialisation at the average of multiple modes and as such a low posterior probability, particularly given the Gaussian approximate posterior we will be adopting \citep{bishopPatternRecognitionMachine2006}. In such circumstances, the mode matching behaviour of the reverse KL may be more appropriate. Computing the reverse KL divergence directly is difficult given our inability to directly evaluate the true log posterior probability. We can circumvent this by appealing to the standard ELBO, evaluated using the reparameterisation trick of \cite{kingma_auto-encoding_2014}, which admits a decomposition consisting of an encoder-independent model evidence term, and the reverse KL we wish to obtain gradients from,
\begin{align}
    \mathcal{L}_{A_R} &= \KL(q(\rvz|\rvx,\pmb\phi) \vert p(\rvz|\rvx)) \\
    &= \E_{q(\rvz|\rvx,\pmb\phi)}\left[\log \frac{q(\rvz|\rvx,\pmb\phi)}{p(\rvz|\rvx)}\right]
\end{align}
where we are once again exclusively interested in obtaining gradients with respect to $\pmb\phi$, and as such, 
\begin{align}
    \!\!\!\!\nabla_{\pmb\phi}\KL(&q(\rvz | \rvx,\pmb\phi) \vert p(\rvz|\rvx)) \nonumber \\
    &= \nabla_{\pmb\phi}\left[ \KL(q(\rvz|\rvx,\pmb\phi) \vert p(\rvz|\rvx)) - \log p(\rvx) \right] \\
    &= \nabla_{\pmb\phi}\left[ -\E_{q(\rvz|\rvx,\pmb\phi)}[\log p(\rvx|\rvz)] \right. \nonumber \\ 
    & \quad\qquad +\left. \KL(q(\rvz;\pmb\phi)\vert p(\rvz)) \right] \\
    &= \nabla_{\pmb\phi} \mathcal{L}_{\text{ELBO}} \label{eq:reverse_kl}
\end{align}

\subsubsection{Jeffrey's Divergence}
By averaging gradients from the forward and reverse KL divergences we may also optimise with respect to (half) the Jeffrey's divergence, also known as the symmetrised KL \citep{jeffreys_invariant_1946}, which can be shown to upper bound 4 times the Jensen-Shannon divergence \citep{lin_divergence_1991}. 
\begin{align}
    \nabla_{\pmb\phi}\mathcal{L}_{A_J} &= \frac{1}{2} \nabla_{\pmb\phi}\left[\KL(p(\rvz|\rvx)\vert q(\rvz|\rvx,\pmb\phi)) \right. \nonumber \\ 
    & \quad + \left. \KL(q(\rvz|\rvx,\pmb\phi) \vert p(\rvz|\rvx))\right]
\end{align}

\subsection{Adaptive Preconditioning}
There now exists a sizeable literature approaching gradient-based sampling from the perspective of optimisation in the space of probability measures \citep{jordan_variational_1998, wibisono_sampling_2018}. This framing has led to the development of analogues to well-known methods from the classical optimisation literature, such as Nesterov's acceleration \citep{ma_is_2019}. Similar analogues to preconditioning have also emerged in the literature, with \citep{girolami_riemann_2011}, demonstrating that an appropriately chosen, possibly position-specific, preconditioning matrix may be used to exploit the natural Riemannian geometry over the induced distributions, improving mixing time and sampling efficiency. A number of works have subsequently capitalized on this technique with a variety of Riemannian metrics, primarily within the context of stochastic gradient Langevin dynamics (SGLD) - a technique that applies Langevin dynamics to noisy mini-batch gradients over deep neural network parameters to obtain posterior samples \citep{welling_bayesian_2011, ahn_bayesian_2012, patterson_stochastic_2013, li_preconditioned_2015}.

Here we adopt the adaptive second-moment computation of the Adam \citep{kingma_adam_2017} optimizer as our preconditioning matrix, computed with iterates over the log unnormalised probability $\log p(\rvx, \rvz_t)$. The resultant algorithm may be considered analogous to the use of the diagonal RMSProp preconditioner for SGLD by \cite{li_preconditioned_2015}, with key differences being in the use of a debiasing step, the use of non-stochastic gradients, and the inclusion of the gradient over the log prior in our second-moment calculations. We note that the Itô SDE associated with an overdamped Langevin diffusion with position-dependent metric tensor $G(\mX_t)$, may be written as \citep{girolami_riemann_2011, ma_complete_2015, roberts_langevin_2002, xifara_langevin_2014}:
\begin{align}
    d\mZ_t = G(\mX_t)\nabla_z \log p(\rvx, \rvz|\pmb\theta)dt &+ \Gamma(\mZ_t)dt \\ 
    &+ \sqrt{2G(\mZ_t)}d\mW_t 
\end{align}
where the term $\Gamma(\mZ_t)$ accounts for changes in local curvature of the manifold, and is defined as:\footnote{We note that this term appears slightly differently to that found in \cite{roberts_langevin_2002} and \cite{girolami_riemann_2011}, as the original formulation was shown by \citep{xifara_langevin_2014} to correspond to the density function with respect to a non-Lebesgue measure (after correcting a transcription error). The term as used in this paper is of the form suggested by \citep{xifara_langevin_2014} which has the required invariant density with respect to the Lebesgue measure.}
\begin{equation}
    \Gamma_i(\mZ_t) = \sum_{j} \frac{G_{ij}(\mZ_t)}{\partial Z_j}
\end{equation}
The resultant discretization given by the Euler-Murayama scheme follows analogously to that in Equation \ref{eq:discretisation}. We follow identically to \cite{ahn_bayesian_2012} and \cite{li_preconditioned_2015} and choose to ignore the $\Gamma_i(\mX_t)$ term in our final discretized algorithm; valid under the assumption that our manifold changes slowly. Our final preconditioned algorithm with amortised warm-starts is described in Algorithm \ref{alg:preconditioned_langevin}. 

\section{Related Works}
Functionally similar algorithms to the one we propose here, have been independently developed from different theoretical perspectives \citep{hoffman_learning_2017, taniguchi_langevin_2022}. Most notably, \citep{hoffman_learning_2017} proposed evolving latent states, also initialized by an inference network, using multiple iterations of Metropolis-adjusted \textit{Hamiltonian Monte Carlo} (HMC) dynamics, with the final state being used to update the parameters of a generative model. 

More recently, \citep{taniguchi_langevin_2022} proposed the application of Metropolis-adjusted Langevin dynamics directly to the parameters of an amortisation network rather than datapoint-wise latent states, leveraging these iterates to also jointly learn a generative model or decoder network. Our work contributes to this growing literature by introducing an approach grounded in computational neuroscience, specifically through the lens of PC. Moreover, to the best of our knowledge, our work is the first to propose and empirically evaluate the use of the Forward KL and Jeffrey's divergence as optimization objectives for a warm-start or inference network, as well as the adaptive preconditioning described herein for improving step size robustness in the unadjusted Langevin algorithm, particularly in the context of learning generative models.

\section{Results}
For all experiments considered here, we adopt generative and warm-start models that are largely coincident with the encoder, and decoder respectively from the VAE architecture of \cite{higgins_beta-vae_2016}, with minor modifications, adopted from more recent VAE models \citep{child_very_2021, vahdat_nvae_2021}, such as SiLU activation functions and softplus parameterised variances. Complete details of model architecture and hyperparameters can be found in Appendix \ref{appendix:experimental_details}
\newcommand{\RightComment}[1]{\hfill $\triangleright$ #1}

\begin{algorithm}[h]
\caption{Preconditioned Langevin PC with Amortized Warm-Starts trained with Jeffrey's Divergence.
For the version corresponding to warm-starts with just the reverse KL, remove the forward KL accumulation and the coefficient of $\frac{1}{2}$ from the reverse KL gradients.
}
\footnotesize
\begin{algorithmic}
\REQUIRE{$\mathcal{D}$: Data-generating distribution}
\REQUIRE{$p(\rvx,\rvz|\pmb\theta)$: Generative model ($\pmb\theta$)}
\REQUIRE{$q(\rvz|\rvx,\pmb\phi)$: Approximate inference model ($\pmb\phi$)}
\REQUIRE{$\beta$: Preconditioning decay rate}
\REQUIRE{$\gamma, \alpha, T$: Langevin step size, parameter learning rate, and number of sampling steps}
\FOR{$\rvx \sim \mathcal{D}$}
\STATE{$\vg_{\pmb\theta}, \vg_{\pmb\phi}, \vm^{(0)}  \gets \pmb0$}
\STATE{$\vz^{(0)} \sim q(\rvz|\vx,\pmb\phi)$} 
\STATE{$\vg_{\pmb\phi} \pluseq \frac{1}{2}\nabla_{\pmb\phi}\mathcal{L}_{A_R}$} \RightComment{Reverse KL gradients}
\FOR{$t \in \{1, 2, \dots, T \}$}
\STATE{$\vg_{\vz} \gets \nabla_{\vz}\log p(\rvx,\vz^{(t-1)}|\pmb\theta)$} \RightComment{Drift}
\STATE{$\vm^{(t)} \gets \beta\cdot\vm^{(t-1)} + (1-\beta)\cdot(\vg_{\vz}^T\vg_{\vz})$} 
\STATE{$\hat{\vm}^{(t)} \gets \sqrt{\vm^{(t)}/(1-\beta^t)}$} \RightComment{Bias correction}
\STATE{$\vz^{(t)} \gets \gamma \cdot \vg_{\vz} \oslash \hat{\vm}^{(t)} + \eta$}, $\eta \sim \mathcal{N}(\pmb0, \text{diag}(2\gamma\cdot\hat{\vm}))$ 
\STATE{$\vg_{\pmb\theta} \pluseq \nabla_{\pmb\theta}\log p(\rvx,\vz^{(t-1)}|\pmb\theta)$} 
\STATE{$\vg_{\pmb\phi} \pluseq \frac{1}{2T}\nabla_{\pmb\phi}\mathcal{L}_{A_F}$} 
\RightComment{Forward KL gradients 
}
\ENDFOR
\STATE{$\pmb\theta \pluseq \alpha \cdot \vg_{\pmb\theta}$} \RightComment{Generative model update}
\STATE{$\pmb\phi \pluseq \alpha \cdot \vg_{\pmb\phi}$} \RightComment{Warm start model update}
\ENDFOR
\end{algorithmic}
\label{alg:preconditioned_langevin}
\end{algorithm}
\subsection{Approximate Inference Objectives}
We begin by investigating the performance of our three approximate inference objectives, the forward KL, reverse KL and Jeffrey's divergence on the quality of our samples when trained with CIFAR-10 \citep{krizhevsky_learning_2009}, SVHN \citep{netzer_reading_2011} and CelebA (64x64) \citep{liu_deep_2015}. As a baseline, we also test with no amortized warm-starts, instead using samples from our prior, for which we adopt an isotropic Gaussian with variance 1, to initialise our Langevin chain. For all tests, we also adopt this prior initialisation for the first 50 batches of training to ameliorate the effects of any poor initialisation in our warm-start models. 

To quantify sample quality we compute the the standard Fréchet distance with Inceptionv3 representations (FID) \citep{heusel_gans_2017} using 50,000 samples.  We observe a largely consistent relationship for the forward KL, with the objective exhibiting both poor performance in terms of sample quality and training instabilities resulting from an increasingly poor initialisation as training progresses. We validate this by recording changes in log probability and the L2 normed gradient of the log probability for random samples during their sampling trajectories for the three objectives. We observe significantly qualitatively different behaviours for the forward KL initialisations, observing drift-dominant conditions with dynamics dominated by maxima-seeking behaviour suggesting poor initialisation far from the mode. Example recordings of the change in log probability $\Delta \log p (\vx, \vz)$ may be found in Figure \ref{fig:delta_log_prob}. Further examples, and the equivalent normed gradient plots can be found in Appendix \ref{appendix:epoch_50_samples}.   

\begin{figure}[!h]
\begin{center}
\includegraphics[width=0.45\textwidth]{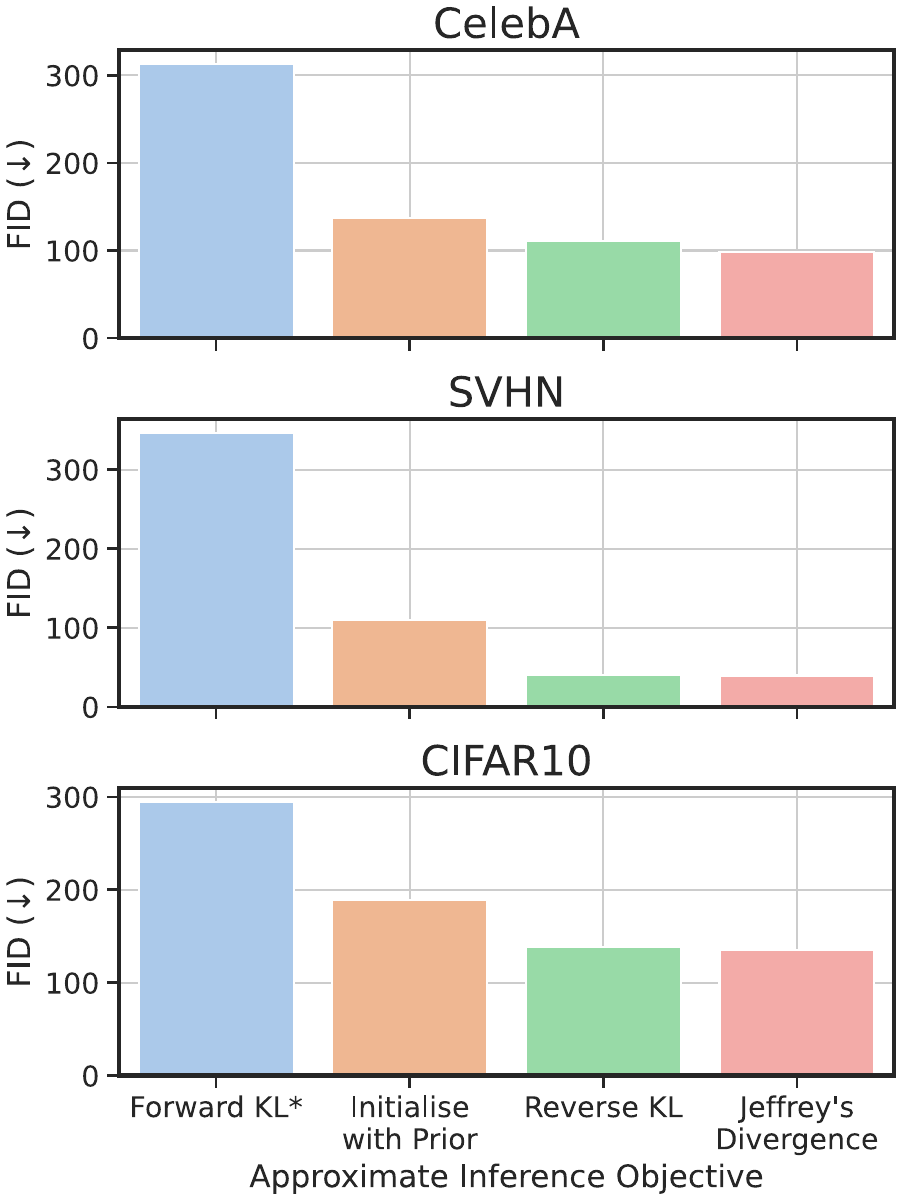}
\end{center}
\caption{FID when using amortised warm-starts trained with our three approximate inference objectives, and baseline with no warm-start model, using initialisation with the prior. $^*$ Values for the forward KL objective are reported for 1 epoch due to the instability of this objective resulting in exploding gradients.}
\label{fig:approx_inference_objectives}

\end{figure}
\begin{figure*}[!h]
\begin{center}
\includegraphics[width=0.95\textwidth]{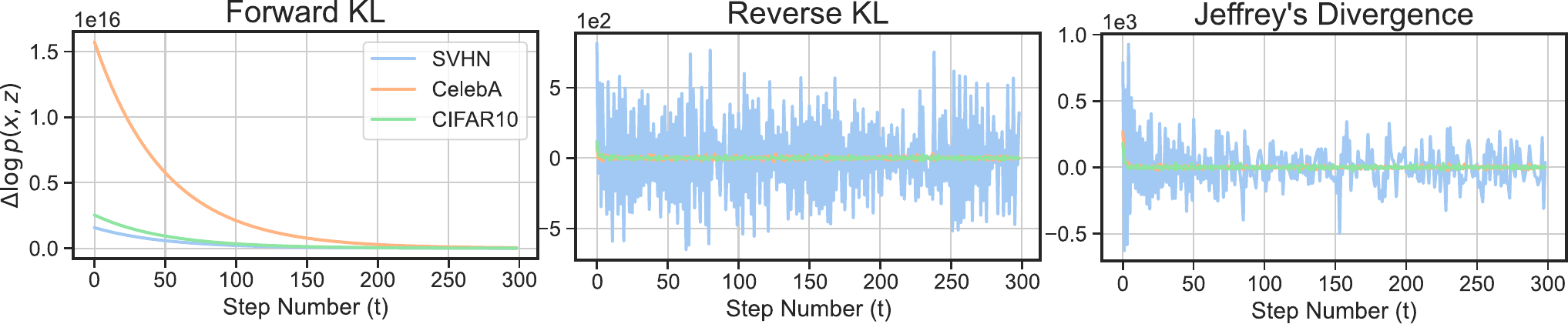}
\end{center}
\caption{Changes in log probability ($\Delta \log p(\vx, \vz)$) during Langevin sampling show forward KL initialisation results in long periods of drift-dominant conditions far from the mode.}
\label{fig:delta_log_prob}
\end{figure*}

In comparison we observe clear improvements in sample quality and FID when using amortised warm-starts trained with Jeffrey's divergence or the reverse KL over the baseline encoder-only models. Due to the improved performance of the Jeffrey's divergence objective in terms of the FID, and qualitatively more diverse sample quality, we adopt this objective for all subsequent experiments. FID values for the three objectives can be found in Figure \ref{fig:approx_inference_objectives}, note that due to exploding gradients for the forward KL objective at later epochs, the FID values in for the forward KL correspond to performance at 1 epoch. 

\subsection{Preconditioning Induced Robustness}

\begin{figure}[!h]
\begin{center}
\includegraphics[width=\textwidth]{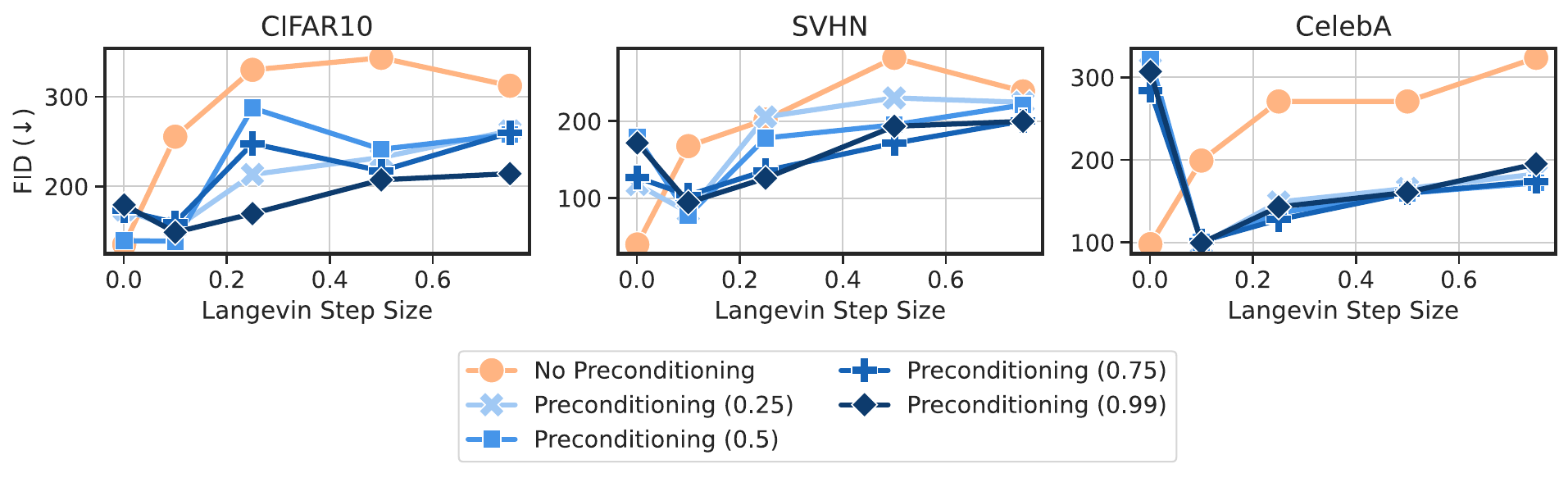}
\end{center}
\caption{FID for Langevin PC models with and without preconditioning across different step-sizes. Numbers in brackets correspond to the preconditioning decay rate ($\beta$). Models trained with preconditioned Langevin dynamics experience significantly less degradation in sample quality at higher step-sizes. With stronger preconditioning generally correlating to the greatest robustness against inference learning rate.}
\label{fig:preconditioning_fid}
\end{figure}

We assess the impact of preconditioning on increasing step sizes by testing models with and without preconditioning as we vary the Langevin step size from 0.001 to 0.5. We observe a substantial protective effect on the degradation of sample quality as step size increases in terms of the FID (Figure \ref{fig:preconditioning_fid}) of the resultant models, with the strength of this protective effect generally correlating with the strength of the preconditioning parameter $\beta$. 

We also find that while preconditioned models exhibit better sample quality over their non-preconditioned counterparts, over the majority of step sizes tested, this trend begins to reverse at the very lowest Langevin step-sizes (1e-3), where non-preconditioned models reach parity or even improved performance. This relationship appears to mirror that of adaptive optimizers for SGD as used in practice, where adaptive optimizers exhibit greater robustness to a wide range of learning rates, but risk being outperformed by standard SGD optimization with a carefully finetuned learning rate. 

\subsection{Samples and Metrics}  
\begin{figure*}[!h]
\begin{center}
\includegraphics[width=\textwidth]{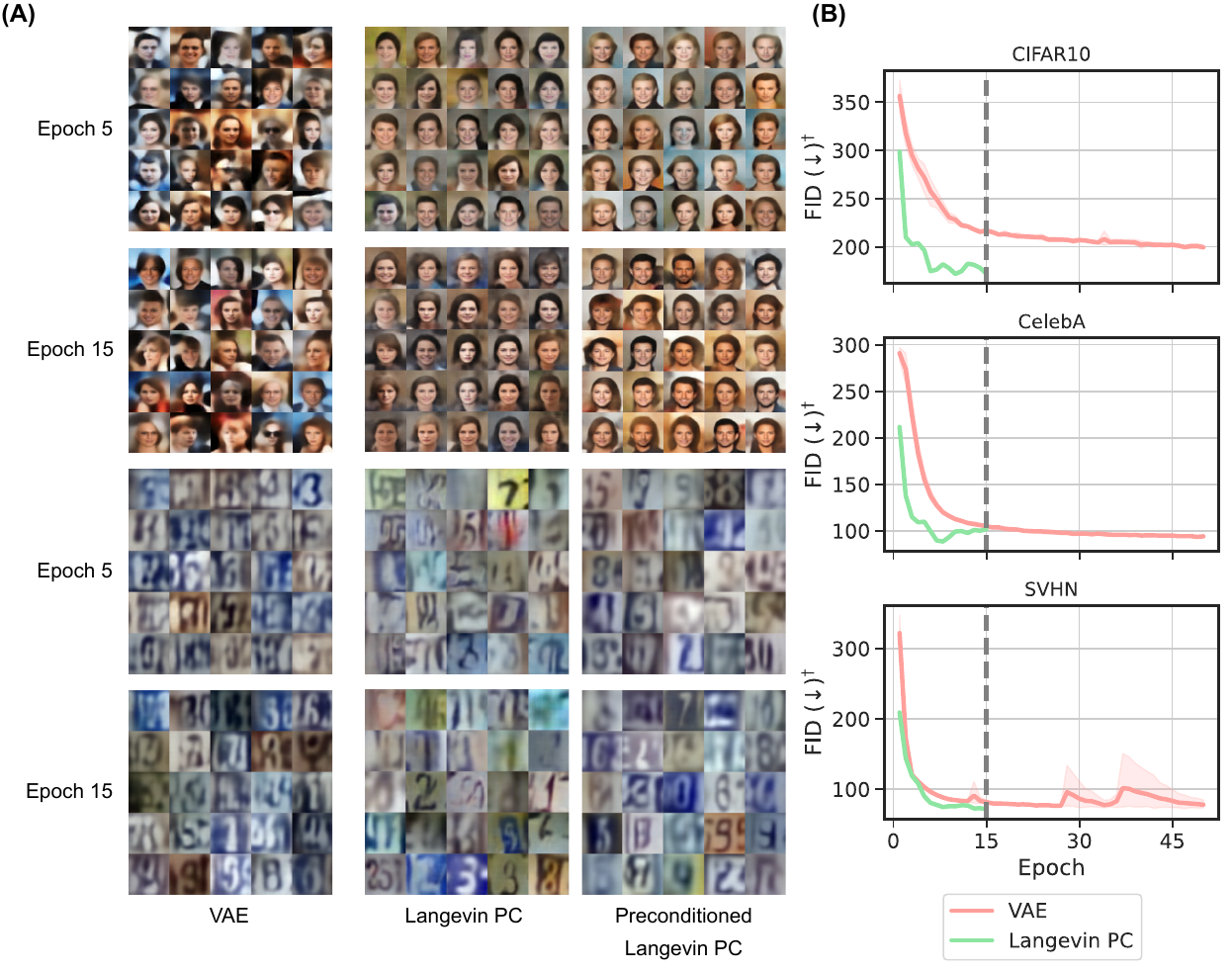}
\end{center}
\caption{
(A) Samples from identical generative models trained as VAEs (left), with LPC (middle), and with preconditioned LPC (right) on CelebA 64x64 (top), and SVHN (bottom). Epoch 50 samples for VAE models can be found in Appendix \ref{appendix:epoch_50_samples}. (B) Sample FID curves of VAE and LPC models throughout training. LPC models generally converge in significantly fewer epochs than their equivalent VAE trained models, with certain models converging in as few as 3 epochs. $^\dag$ Note: FID values reported in this graph are calculated online during training using significantly fewer samples than the post-training values reported in Table \ref{tab:final_results}, and may thus differ in precise value.  
}
\label{fig:samples_and_convergence}
\end{figure*}

We trained identical generative models using the standard VAE objective, alongside the LPC methodologies described herein. VAE models were hyperparameter tuned on learning rates with the best performing model with respect to FID being chosen for comparison. LPC models were analogously tuned on inference learning rate and preconditioning strength. Remaining hyperparameters were kept constant between runs such as optimizer, batch size and prior variance, to ensure a fair and like-for-like comparison. Full experimental details may be found in Appendix \ref{appendix:experimental_details}, alongside the optimal hyperparameters selected for each dataset. 

LPC and VAE models were trained for 15 and 50 epochs respectively. To evaluate sample quality we computed FID (using 50,000 samples), as well as density and coverage \citep{naeem_reliable_2020} - a more robust alternative to precision and recall metrics - also using Inceptionv3 embeddings.  

LPC models demonstrated comparative or better performance to their VAE counterparts. In particular, LPC models out-performed VAE models trained for more than 3 times as many SGD iterations (50 epochs vs 15), on SVHN and CIFAR10, in terms of FID, as well as on CelebA and CIFAR10 with respect to density and coverage. Samples from LPC models were also markedly less blurry - or more sharp - in comparison to VAE counterparts, an issue known to plague VAE models. (See Figure \ref{fig:samples_and_convergence}, for some non-cherry picked examples). 

\begin{table}[!h]
\centering
\caption{Comparative evaluation of FID, Density, and Coverage for LPC and VAE models across different datasets.}
\vspace{10pt} 
\label{tab:final_results}
\begin{tabular}{
  c
  S[detect-weight, table-format=3.2]
  S[detect-weight, table-format=3.2]
  S[detect-weight, table-format=3.2]
}
\toprule
Dataset & {FID (↓)} & {Density (↑)} & {Coverage (↑)} \\
\midrule
\multicolumn{4}{c}{LPC (15 Epochs)} \\
\midrule
CelebA   & 97.49 & \bfseries 0.54 & \bfseries 0.13 \\
SVHN     & \bfseries 39.64 & 0.33 & 0.42 \\
CIFAR10  & \bfseries 113.29 & \bfseries 0.63 & \bfseries 0.13 \\
\midrule
\multicolumn{4}{c}{VAE (15 Epochs)} \\
\midrule
CelebA   & 90.63 & 0.10 & 0.08 \\
SVHN     & 53.88 & 0.60 & 0.39 \\
CIFAR10  & 183.21 & 0.06 & 0.03 \\
\midrule
\multicolumn{4}{c}{VAE (50 Epochs)} \\
\midrule
CelebA   & \bfseries 82.09 & 0.16 & 0.12 \\
SVHN     & 44.76 & \bfseries 0.65 & \bfseries 0.48 \\
CIFAR10  & 145.87 & 0.14 & 0.06 \\
\bottomrule
\end{tabular}
\end{table}
\section{Discussion}
\label{sec:conclusion}

We have presented an algorithm for training generic deep generative models that builds upon the PC framework of computational neuroscience and consists of three primary components: an unadjusted overdamped Langevin sampling, an amortised warm-start model, and an optional light-weight diagonal preconditioning. We have evaluated three different objectives for training our amortised warm-start model: the forward KL, reverse KL and the Jeffrey's divergence, and found consistent improvements when using the reverse KL and Jeffrey's divergence over baselines with no warm-starts (Figure \ref{fig:approx_inference_objectives}). We have also evaluated our proposed form of adaptive preconditioning and observed an increased robustness to increaing Langevin step size (Figure \ref{fig:preconditioning_fid}). Finally, we have evaluated the resultant Langevin PC algorithm by training like-for-like models with the standard VAE methodology or the proposed Langevin PC algorithm. We have observed comparative or improved performance in a number of key metrics including sample quality, diversity and coverage (Table \ref{tab:final_results}), while observing training convergence in a fraction of the number of SGD training iterations (Figure \ref{fig:samples_and_convergence}B).

\subsection{Future directions}

Langevin predictive coding opens doors in two different directions. The first is in regards to PC as an instantiation of the Bayesian brain hypothesis and as a candidate computational theory of cortical dynamics. In this setting, the introduction of Gaussian noise into the PC framework may represent more than simply an implementational detail associated with Langevin sampling but rather a deeper phenomena rooted in the ability of biological learning systems such as the brain to utilise sources of endogenous noise to their advantage. 

It is well known that neuronal systems, including their dynamics and responses, are rife with noise at multiple levels \citep{faisal_noise_2008, shadlen_variable_1998}. These sources of noise arise from, amongst other things, stochastic processes occuring at the sub-cellular level, impacting neuronal response through, for example, fluctuations in membrane-potential \citep{derksen_fluctuations_1966}. Yet the precise role of such randomness, in information processing, continues to be an open question \citep{mcdonnell_benefits_2011, deco_resting_2013}. The Langevin PC algorithm suggests one such role may be in the principled exploration of the latent space of hypotheses under one's generative model. 

Secondly, from the perspective of Langevin PC as an in-silico generative modelling algorithm we note a number of interesting avenues that we have not had the time to explore here. 
These include:
\begin{itemize}
    \item Models with a hierarchy of stochastic variables, such as those found in most state of the art VAE models \citep{child_very_2021, vahdat_nvae_2021, hazami_efficient-vdvae_2022}. Which may require adopting a corresponding top-down hierarchical warm-start model. 
    \item Automatic convergence criteria for determining when our Markov chain has converged to a certain level of error \citep{roy_convergence_2020}.
    \item Underdamped Langevin dynamics, which incorporate auxiliary momentum variables into the Langevin sampling to achieve an accelerated rate of convergence \citep{cheng_underdamped_2018, ma_is_2019}.
\end{itemize}

\subsection{Limitations}
The methods we propose here are not without limitation. When implemented on current in-silico autograd frameworks, the need to enact multiple sequential iterations of Langevin dynamics for each SGD iteration requires additional computational cost and thus wall-clock time. In practice, this additional wall-clock time is counteracted, to an extent, by the increased efficiency of the Langevin PC algorithm in terms of the number of SGD iterations required to obtain similar or better performance as their VAE counterparts. When accounted for, we nonetheless observed end-to-end wall clock times for training that were approximately x7 and x11 slower for LPC algorithms using the reverse KL and Jeffrey's divergence respectively. (See Appendix \ref{appendix:wall_clock_time} for per batch timings and relative slow downs). 

We note that this additional cost is isolated to training, whereas the cost of sampling LPC models remain equivalent to their VAE counterparts - requiring a single ancestral sample, or forward evaluation, through the generative model to obtain. For models deployed for long-term use, such inference costs account for the bulk of computational cost. Therefore, LPC may be a viable candidate to improve model quality without increasing inference cost when deployed. We also speculate that the form of these dynamics - precision-weighted prediction errors with additive Gaussian noise - may render them a good candidate for implementation on analog hardware, where such dynamics would be enacted by the intrinsic but noisy fast-timescale physics of such systems. 

\bibliography{references}
\bibliographystyle{iclr2024_conference}

\newpage
\appendix
\section{Appendix}

\subsection{Experimental Details}
\label{appendix:experimental_details}

All experiments in this paper adopted the following network architectures for the generative model and approximate inference models. These models are derived from the encoder/decoder VAE architectures of \citep{higgins_beta-vae_2016} with slight modifications such as the use of the SiLU activation function adopted in more recent VAE models such as \citep{hazami_efficient-vdvae_2022, vahdat_nvae_2021}.

\begin{table}[!h]
    \centering
    \begin{tblr}{
      colspec = {@{}Q[c,m]Q[c,m]@{}}, vlines, hlines
    }
         Generative Model ($\log p(\rvx, \rvz|\pmb\theta)$ & Warm-Start/Encoder Model ($\log q(\rvz|\rvx, \phi)$)  \\ \hline
         Latent Dim = 40 & Obs Dim = (64, 64) or (32, 32) or (28, 28) \\ 
         Linear(256) & {If Input = (64,64): Conv(32, 3, 3, 1) \\ else: Conv(32)} \\ 
         SiLU & SiLU \\
         Conv(64, 4, 1, 0) & Conv(32) \\
         SiLU & SiLU \\
         Conv(64) & Conv(64) \\
         SiLU & SiLU \\
         Conv(32) & {If Obs Dim = (28, 28): Conv(64, 3) \\ else: Conv(64)}\\
         SiLU & SiLU \\
         {If obs dim = (64, 64): Conv(32) \\ else if obs dim = (28, 28): Conv(32, 3, 1, 0) \\ else: Conv(32, 3, 1, 1)} & Conv(256, 4) \\
         SiLU & SiLU \\
         Conv(3) & {Linear(2*40) \\ (Softplus(beta=0.3) applied to variance component)} 
    \end{tblr}
    \caption{Layer argument definitions are Conv(Number of out channels, kernel size, stride, padding), and Linear(Output dimensions) for 2d convolution and linear layers respectively. Kernel size, stride and padding are 4x4, 2, and 1 respectively if not explicitly stated.}
    \label{tab:model_architecture_details}
\end{table}

\begin{table}[!h]
    \centering
    \begin{tblr}{
      colspec = {@{}X[c]X[c]@{}}, vlines, hlines
    }
         Hyperparameter & Value \\ \hline
         Optimizer & Adam \\
         Learning Rate ($\alpha$) & 1e-3 \\
         Batch size & 64 \\
         Output Likelihood & Discretised Gaussian \\
         Max Sampling Steps ($T$) & 300 \\
         Preconditioning Decay Rate ($\beta$) & 0.99
    \end{tblr}
    \caption{Default hyperparameters used in experiments unless explicitly stated. Note: some of these are varied as part of ablation tests, see main text for more details. }
    \label{tab:hyperparameters}
\end{table}

Optimal learning rates for VAE were found to be 1e-3, 8e-4 and 1e-3 for CIFAR10, CelebA and SVHN respectively. For LPC, optimal inference learning rates were found to be 1e-1, 1e-1, and 1e-3 with $\beta$ equal to 0.25, 0.25 and 0 (No preconditioning), for CIFAR10, CelebA and SVHN respectively. 

\subsection{Low-Dimensional Projection of Inference and Sampling Trajectories}
\label{appendix:pca_latent_trajectory}

The problem of visualising high-dimensional trajectories is a well-known one which generally arises in the context of visualising the stochastic gradient descent trajectories of high-dimensional weights in neural networks \citep{gallagher_visualization_2003, li_visualizing_2017, lipton_stuck_2016}. 

Here we adapt the method suggested by \citep{li_visualizing_2017} to visualise the inference or sampling trajectories of our latent states $\rvz^{(t)}$. We apply principle component analysis (PCA) to the series of vectors pointing from our final state to our intermediate states, i.e. $[\rvz^{(1)}-\rvz^{(T)}, \dots, \rvz^{(T-1)}-\rvz^{(T)}]$, and project our trajectories on the first two principle components. We visualise the projected trajectories on top of the loss landscape of the negative potential (log joint probability) by evaluating our generative model across a grid of latent states linearly interpolated in the direction of the principle components around the final state. 

\begin{figure}[h]
\begin{center}
\includegraphics[width=\textwidth]{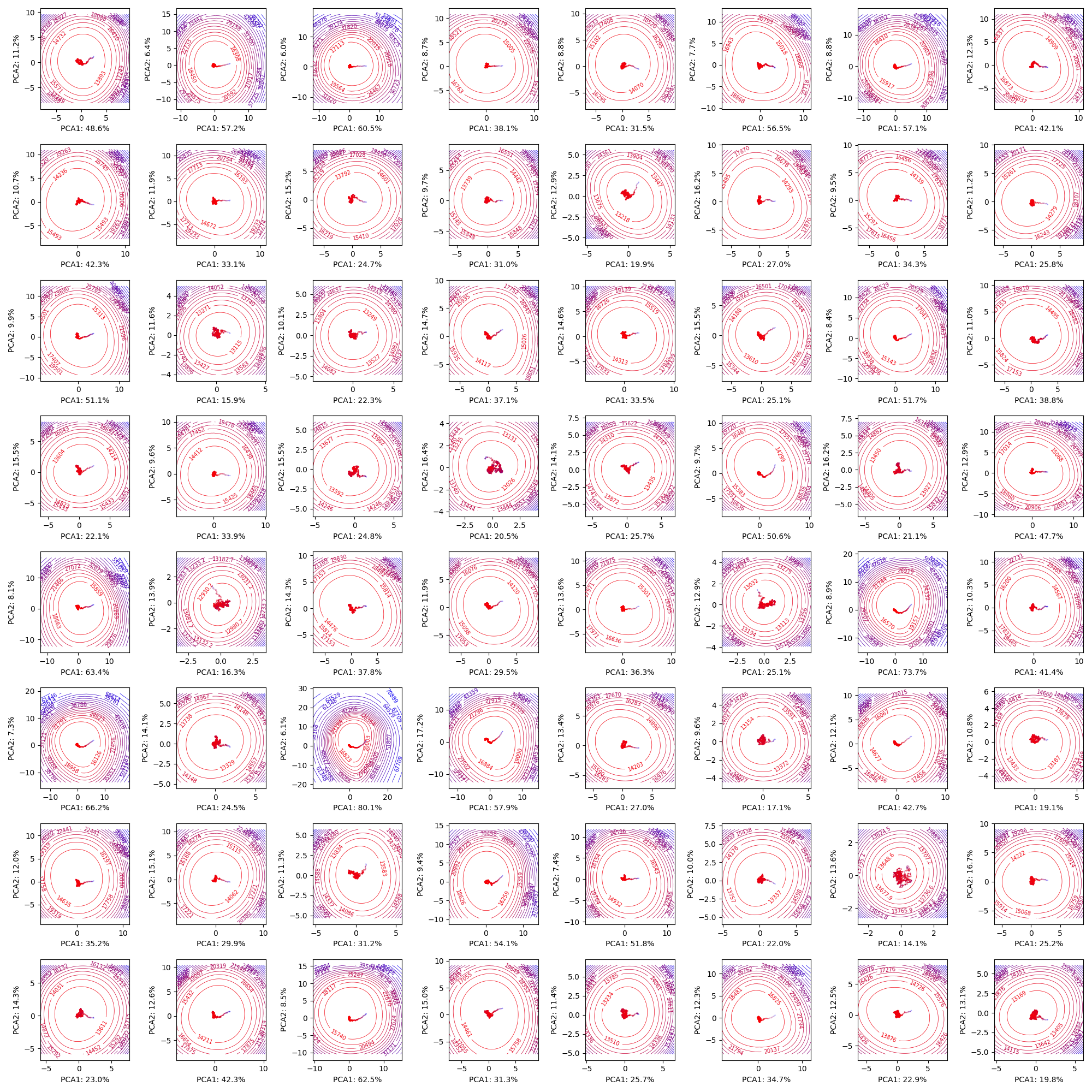}
\end{center}
\caption{Projection of a 64 sample batched high-dimensional latent state trajectories under Langevin PC sampling. Contour lines and hue correspond to values of the negative log joint probability (blue high, red low), marker brightness corresponds to time-step (earlier is lighter).} 
\label{fig:batched_projection_1}
\end{figure}

Projections of an example batch of sampling trajectories can be seen in Figure \ref{fig:batched_projection_1}. 

\clearpage
\subsection{Additional Samples and Figures}
\label{appendix:epoch_50_samples}

\begin{figure}[!h]
\begin{center}
\includegraphics[width=\textwidth]{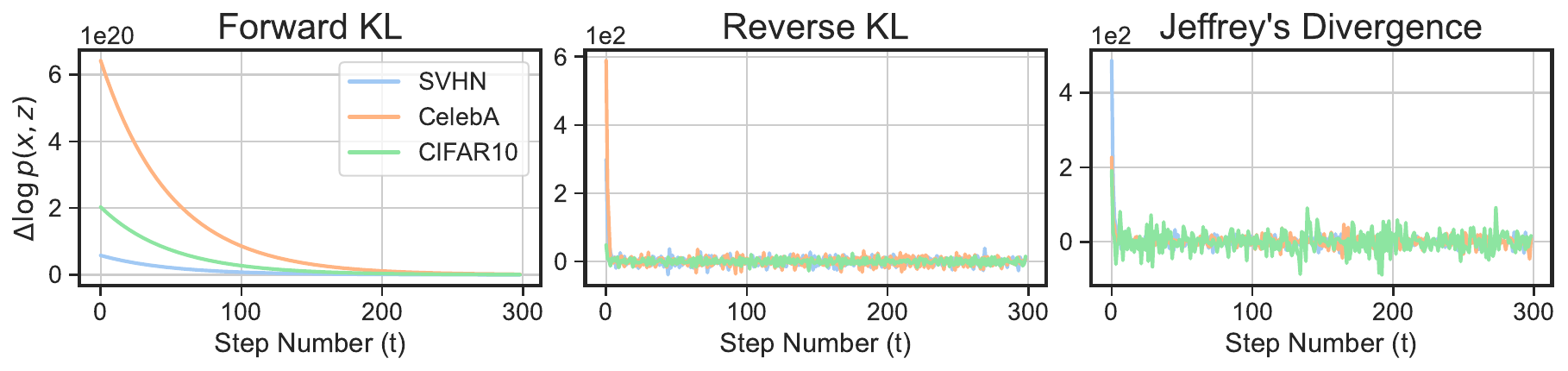}
\end{center}
\caption{Log probability changes during Langevin sampling for samples from training batch 600 for our three approximate inference objectives. }
\label{fig:additional_delta_log_prob}
\end{figure}

\begin{figure}[!h]
\begin{center}
\includegraphics[width=\textwidth]{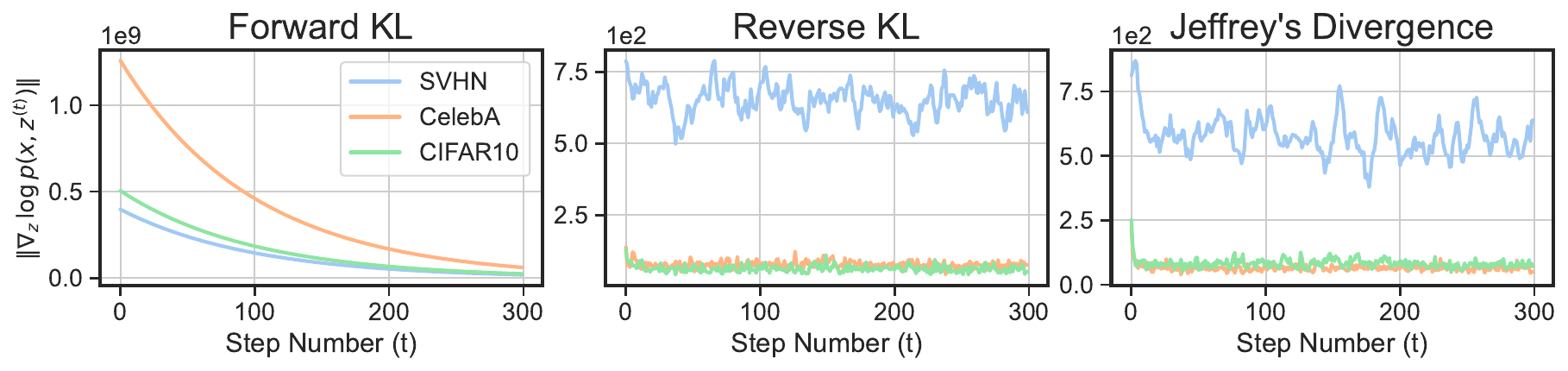}
\includegraphics[width=\textwidth]{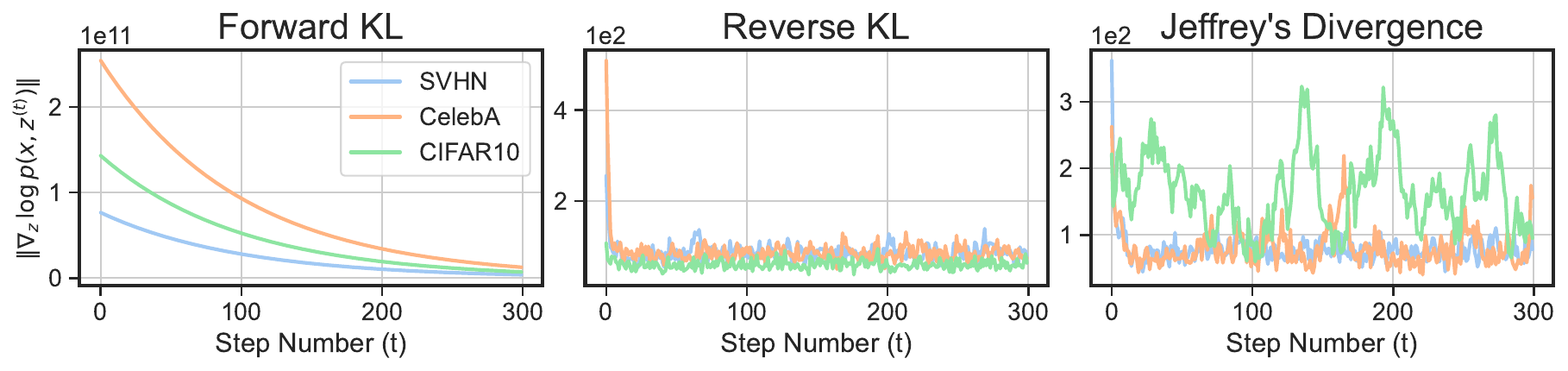}
\end{center}
\caption{L2 normed log probability during Langevin sampling for samples from training batch 300 (top) and 600 (bottom), for our three approximate inference objectives. }
\label{fig:normed_grad}
\end{figure}

\begin{figure}[!h]
\begin{center}
\includegraphics[width=0.2\textwidth]{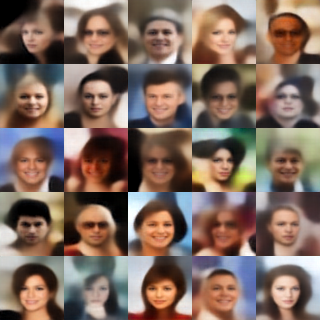}
\includegraphics[width=0.2\textwidth]{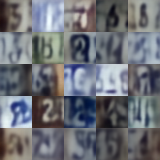}
\end{center}
\caption{Epoch 50 samples from VAEs trained on CelebA 64x64 (left), and SVHN (right)}
\label{fig:epoch_50_samples}
\end{figure}

\clearpage
\subsection{Wall Clock Time}
\label{appendix:wall_clock_time}
\begin{table}[!h]
\centering
\caption{Batch times and end to end slowdowns for LPC algorithms as recorded on an Nvidia 4090. End to end refers to 15 epochs for LPC algorithms, and 50 epochs for VAE algorithms.}
\vspace{10pt}
\label{tab:lpc_slowdown}
\begin{tabular}{lccc}
\toprule
Algorithm      & Per batch time (ms) & {Per batch slowdown} & {End to end slowdown} \\
\midrule
VAE & 0.022 & x1 & x1 \\
LPC (Reverse)  & 0.533 & x24                  & x7                    \\
LPC (Jeffreys)  & 0.798 & x36                  & x11                   \\
\bottomrule
\end{tabular}
\end{table}
\end{document}